\documentclass[letterpaper, 10 pt, conference]{ieeeconf}  

\usepackage{hyperref}
\hypersetup{
    colorlinks=true,
    linkcolor=blue,
    filecolor=magenta,      
    urlcolor=cyan,
    pdftitle={cramp},
    pdfpagemode=FullScreen,
    }
\IEEEoverridecommandlockouts                              

\usepackage{hyperref}

\usepackage{amsmath, amstext} 
\usepackage{amssymb}  
\usepackage{bm}

\usepackage{algorithm}
\usepackage{algpseudocode}
\usepackage{setspace}

\usepackage{pgfplots}
\pgfplotsset{compat=newest} 

\usepackage{multirow}
\usepackage{booktabs} 
\usepackage{makecell} 
\usepackage{caption}
\newcommand{\vs}{\bm{s}}
\newcommand{\vtau}{\bm{\tau}}
\newcommand{\vepsilon}{\bm{\epsilon}}

\usepackage{xcolor}

\title{\LARGE \bf
Adaptive Planning with Generative Models under Uncertainty
}

\author{Pascal Jutras-Dubé, Ruqi Zhang, and Aniket Bera \\
{\textit{Department of Computer Science, Purdue University, USA}} 
\\\textbf{Implemention}: \href{https://pascaljd.github.io/ensemble-adaptive-policy/}{https://pascaljd.github.io/ensemble-adaptive-policy/}
}

\begin{document}

\maketitle
\thispagestyle{empty}
\pagestyle{empty}

\begin{abstract}
Planning with generative models has emerged as an effective decision-making paradigm across a wide range of domains, including reinforcement learning and autonomous navigation.
While continuous replanning at each timestep might seem intuitive because it allows decisions to be made based on the most recent environmental observations, it results in substantial computational challenges, primarily due to the complexity of the generative model's underlying deep learning architecture.
Our work addresses this challenge by introducing a simple adaptive planning policy that leverages the generative model's ability to predict long-horizon state trajectories, enabling the execution of multiple actions consecutively without the need for immediate replanning.
We propose to use the predictive uncertainty derived from a Deep Ensemble of inverse dynamics models to dynamically adjust the intervals between planning sessions.
In our experiments conducted on locomotion tasks within the OpenAI Gym framework, we demonstrate that our adaptive planning policy allows for a reduction in replanning frequency to only about 10\% of the steps without compromising the performance. 
Our results underscore the potential of generative modeling as an efficient and effective tool for decision-making.
\end{abstract}

\section{Introduction}

In recent years, the domain of generative modeling has witnessed transformative advancements, marked by the development of image synthesis models like DALL-E \cite{ramesh2022dalle} and Stable Diffusion \cite{rombach2022stablediffusion}. This technological progression has extended to the generation of high-quality videos from text prompts \cite{ho2022imagen, openai2024sora}. Concurrently, language models like GPT \cite{brown2020GPT} have achieved significant milestones in generating coherent text and engaging in conversations based on brief text prompts.

Recently, generative models have been applied to offline reinforcement learning (RL), where the goal is to derive optimal policies from previously collected datasets. 
The challenge of predicting future states and actions can be formulated as a sequence modeling task, which can be addressed through generative modeling \cite{janner2021sequence, chen2021decisiontransformer, ajay2022decisiondiffuser}.

However, the state prediction process 
incurs substantial computational costs due to the deep neural network architecture of the generative models  \cite{janner2022diffuser, ajay2022decisiondiffuser}.
These computational demands can be a problem in real-time decision-making applications, where agents must rapidly take an action within a time-constrained control loop to plan or adjust their trajectory in response to new environmental observations.

Efforts to improve the sampling efficiency of generative models form a substantial body of work, but few strategies have been specifically developed for decision-making contexts.  
Most solutions are tailored to the specific architectural features of the generative models they use~\cite{janner2022diffuser, zhou2024adaptive}.
Such model-specific methods, while effective, are constrained by their limited applicability across different models. 

\begin{figure*}[tbh!]
  \centering
  \includegraphics[width=\textwidth]{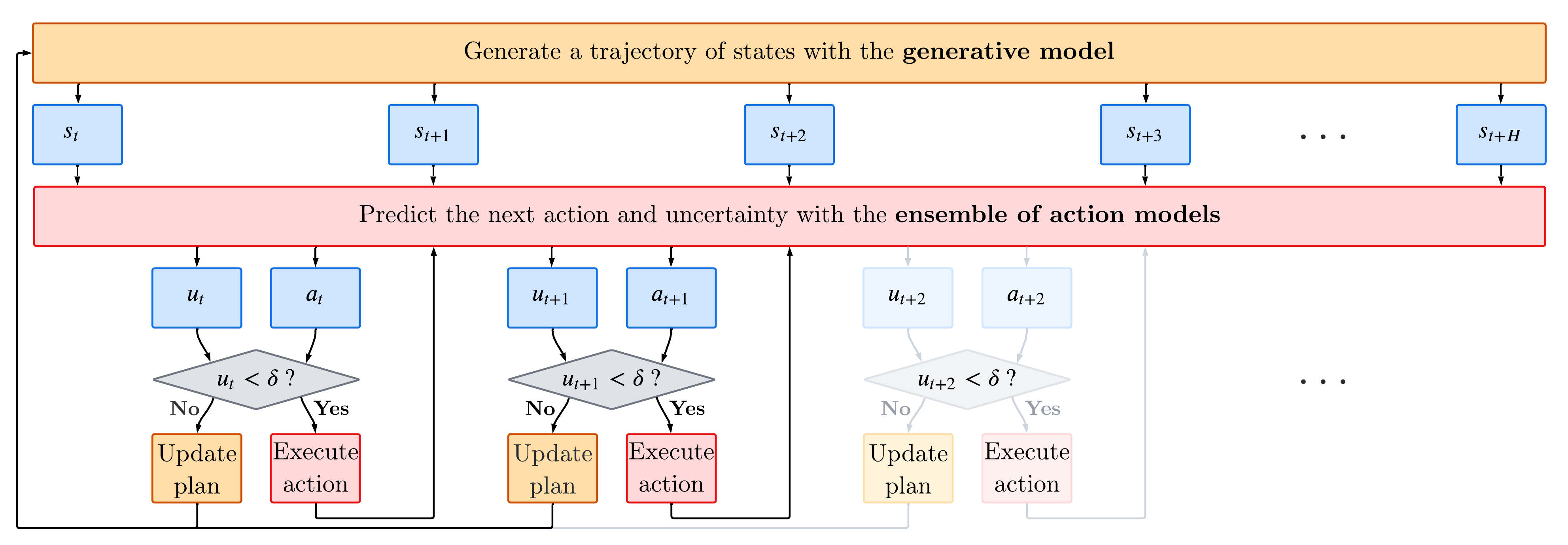}
  \caption{\textit{The generative model generates a trajectory of states and the action model computes the initial action. The policy continuously predicts and executes subsequent actions as long as the uncertainty remains below a predefined threshold.}} 
  \label{fig:adaptive_policy}
\end{figure*}

In this work, we introduce a novel approach that leverages the inherent structure of the decision-making problem to enhance the efficiency of the control process.
We use a generative model to predict a trajectory of future environmental states, and a much smaller action model to determine the next actions based on this trajectory. 
Although planning with generative models is computationally intensive, it enables the prediction of long horizons of future states. 
Drawing on this observation, our approach executes multiple actions consecutively, thereby reducing the frequency of calls to the generative model. 
To determine the optimal times for updating the plan and invoking the generative model anew, we use the uncertainty in the action model’s predictions as a guiding criterion.
The proposed adaptive policy is illustrated in Figure \ref{fig:adaptive_policy}.

Our work introduces the following contributions.
\begin{itemize}
    \item We introduce a simple adaptive policy that enhances the planning process with generative models by leveraging the action model's confidence levels, enabling faster decision-making. 
    Unlike existing solutions, 
    our approach can be universally applied to different generative models without requiring any alterations.
    Our approach utilizes Deep Ensembles \cite{lakshminarayanan2017deepensembles} for efficient and effective predictive uncertainty estimation, allowing for dynamic adjustments in the planning based on the model's confidence.
    \item We conduct a comprehensive evaluation of our method using the D4RL benchmark \cite{fu2020d4rl}.
    Our results demonstrate that our approach can achieve planning speeds of more then 50 times faster than the prior art. 
    This improvement in speed is achieved with no or minimal impact on the rewards. 
\end{itemize}

The remainder of this paper is structured as follows: Section~\ref{sec:relatedwork} reviews related work, emphasizing generative modeling in decision-making and efforts to enhance the sampling speed, with a focus on diffusion models. 
Section~\ref{sec:background} conceptualizes offline RL as a sequence modeling task, detailing how the problem can be split into two stages: generating a sequence of states over a long horizon, followed by action prediction. 
Section \ref{sec:adaptive} introduces our novel adaptive policy that leverages Deep Ensembles. 
Section \ref{sec:experiments} examines the empirical performance of our adaptive policy, highlighting its effective balance between speed and efficacy. 
The paper concludes with Section \ref{sec:discusionconclusion}, where we summarize our findings, discuss limitations, and outline directions for future research.

\section{Related Work}\label{sec:relatedwork}

\subsection{Generative Modeling for Decision-Making}
Janner et al. \cite{janner2021sequence} and Chen et al. \cite{chen2021decisiontransformer} reimagine reinforcement learning (RL) as a sequence modeling problem, deviating from the traditional approach that relies on estimating policies based on the Markov property. 
They implement a Transformer \cite{Vaswani2017transformeer} to model distributions over states trajectories. 
This represents a conceptual shift away from conventional reinforcement learning techniques, which primarily concentrate on estimating value functions or determining policy gradients.

Diffusion models \cite{sohl2015thermo, song2019smld, ho2020ddpm, song2021sde} progressively perturb data towards noise via a Gaussian process and generate data by reversing that process.
Diffuser \cite{janner2022diffuser} is a diffusion model designed for planning. 
It differs from traditional model-based planning by predicting entire trajectories, enhancing scalability for long horizon planning. 
While Diffuser diffuses over both states and actions, Decision Diffuser \cite{ajay2022decisiondiffuser} diffuses only across states and models the action space with an inverse dynamics model. 
The choice to diffuse only over states is justified by the inherent challenges of modeling actions, which often represent complex, discrete, or force-based decisions.
Other recent works have explored the use of diffusion models for planning, underscoring the potential of diffusion-based approaches in decision-making \cite{wang2023diffusionpolicy, chi2023visuomontor, liang2023adaptdiffuser, he2024diffusion}.

\subsection{Improving Sampling Speed}
The iterative refinement process of diffusion models is computationally expensive, typically requiring tens to hundreds of calls to the underlying deep neural network. Consequently, improving the sampling speed of diffusion models has become an intensive domain of research. 
Recent works \cite{jolicoeurmartineau2021gotta, karras2022edm} have demonstrated that using 2nd order stochastic differential equation solvers for denoising offers an excellent balance between sample quality and network evaluations, achieving impressive results in image generation with as few as 36 network function evaluations. 
Additionally, knowledge distillation techniques have significantly accelerated the sampling speed of diffusion models~\cite{luhman2021knowledge, salimans2022progressive, meng2023guidedprogdist, song2023consistency}. 
Moreover, a new class of generative models known as consistency models \cite{song2023consistency, song2023improved} are designed to overcome the low sampling speed inherent in diffusion models by directly mapping any point on the noise trajectory to the data space.
This approach has shown promising results in image generation with as few as two and even one-step generation. 

In planning, similar to sampling with consistency models, Janner et al. \cite{janner2022diffuser} propose warm-starting the generative process by adding partial noise to the previously generated trajectory and running the corresponding number of denoising steps.
Concurrent with our work, Replanning with Diffusion Models \cite{zhou2024adaptive} assesses when to replan based on the likelihood of existing plans. Their approach calculates trajectory feasibility by introducing Gaussian noise and evaluating the KL divergence during denoising, a process that is distinctly model-specific.
The accuracy of the likelihood estimation is directly affected by how well the diffusion model has learned the distribution of successful trajectories. 
This means that any limitations in the diffusion model's training data or its capacity to capture the complexity of the environment could affect the reliability of the replanning criterion.
    
In our work, we leverage the inherent long-horizon prediction of generative models to execute multiple actions in a row, and use uncertainty associated with the action model's predictions as a criterion on when to resample.

\subsection{Estimating Uncertainty in Neural Networks}
Traditionally, quantifying uncertainty leans on a Bayesian framework, where a prior distribution is defined over the network's parameters. 
Given the training data, the posterior distribution over the parameters is computed, which is used to quantify predictive uncertainty. 
However, due to the intractability of Bayesian inference in neural networks, various approximation methods have been proposed.

Markov Chain Monte Carlo (MCMC) methods  \cite{neal1996bayesian} approximate sampling from the posterior distribution by constructing a Markov chain that has the desired distribution as its equilibrium distribution. 
However, MCMC is computationally expensive because it requires a large number of steps to achieve convergence, making it challenging for large-scale or real-time applications.
Variational inference techniques \cite{blundell2015weight, graves2011practical} provide a scalable alternative to MCMC by approximating the posterior distribution with a simpler, parameterized distribution. 
These techniques optimize the parameters of the simpler distribution to minimize the difference between the true posterior and the approximation, often measured by the Kullback-Leibler divergence.

Monte Carlo dropout \cite{gal2016dropout} stands out for its simplicity in approximating Bayesian inference. 
By incorporating dropout during both training and inference, this method simulates sampling from the network's posterior distribution. 
The uncertainty estimation process entails multiple forward passes with dropout, culminating in an ensemble of outputs. 
The aggregation of these outputs yields an estimate of the predictive mean and variance, providing insights into the network's uncertainty regarding its predictions.

Deep ensembles \cite{lakshminarayanan2017deepensembles} further simplify the uncertainty estimation. 
This method involves training several network instances from different initializations and combining their predictions. 
Such an approach not only captures the inherent and model-specific uncertainties but also does so without necessitating intricate changes to the network architecture or its training protocol.

\section{Background}\label{sec:background}

\subsection{Problem Description}
Let $\mathcal S$ and $\mathcal A$ be the state and action spaces, respectively, and  $\mathcal R: \mathcal S \times \mathcal A \mapsto \mathbb R$ be a reward function.
A trajectory is a sequence of $T$ states, actions, and rewards 
\begin{equation}\label{eq:agent}
    \vtau = (s, a, r)_t \in \mathcal S \times \mathcal A \times \mathbb R, 0 \leq t < T,
\end{equation} 
and its return is the sum of the time-steps rewards $R(\vtau)=\sum_{t}r_t$.
The goal of the agent is to learn a policy $\pi : \mathcal S \mapsto \mathcal A$ that predicts the next action to take given the current environment's state, such that it maximizes the expected return $\mathbb E \left[ R(\vtau) \right]$ over the trajectories.

In offline reinforcement learning, the agent learns from a dataset of trajectories that were collected through various, potentially suboptimal, policies. 
Unlike in online reinforcement learning, the agent does not have the opportunity to explore the environment or collect new data based on its current policy. 

At test time, the environment is initialized with a state randomly selected from an initial state distribution, $s_0 \sim \rho_0$. 
Each action results in a new state, determined by the state transition function $\mathcal{T}: \mathcal{S} \times \mathcal{A} \mapsto \mathcal{S}$. 
This process repeats in a receding horizon control loop until a termination condition is met, ending the episode.

The task of policy learning can be divided into two phases: sequence modeling of future states followed by action generation based on those predictions.


\subsection{Generative Modeling for States Prediction}\label{sec:decisiondiffuser}
The first phase involves learning the data distribution of state trajectories. 
Given a training dataset of trajectories $\mathcal D = \{\vtau^i\}_{0 \leq i < N}$, we extract sequences of states $\vs = (s)_j, 0\leq j < H$ of horizon $H$.
The goal is to estimate the underlying conditional distribution $p_{\text{data}}(\cdot|s_t)$, where $s_t$ is the initial state from which the prediction of future states begins.
We create a model that represents a parameterized distribution $p_\theta$, and we tune its parameters, $\theta$, by minimizing a divergence between $p_\theta$ and $p_{\text{data}}$.
Then we can generate new sequences of states by sampling from the model distribution $\hat \vs = (s_t, \hat s_{t+1}, \ldots, \hat s_{t+H-1}) \sim p_{\theta}(\vs|s_t)$.
We use the hat notation to distinguish between predicted and observed states.


Following Janner et al. \cite{janner2022diffuser}, we use a Denoising Diffusion Probabilistic Model (DDPM) \cite{ho2020ddpm} to generate state sequences. 
DDPM progressively adds noise to a data distribution until it becomes pure noise, and subsequently reverse that process through a Markov chain with learned transition kernels to generate trajectories from noise.

Given the data distribution $\vs^0 \sim p_{\text{data}}(\vs^0)$, the forward noising process produces a sequence of random vectors $\vs^1, \vs^2, \ldots, \vs^K$ with the transition kernel 
\begin{equation*}
    q(\vs^{k+1}|\vs^k) = \mathcal N(\vs^{k+1}; \sqrt{\alpha_k}\vs^k, (1 - \alpha_k)\bf I)
\end{equation*}
where $\alpha_k$ is the noise scale schedule.
The number of diffusion steps $K$ is chosen big enough such that $\vs^K$ approximately follows a standard Normal distribution.
The reverse denoising process is modeled by the learnable transition kernel 
\begin{equation*}
    p_\theta(\vs^{k-1}|\vs^k) = \mathcal N (\vs^{k-1}|\mu_\theta(\vs^k, k), \Sigma_k),
\end{equation*}
starting with $\vs^K \sim \mathcal N(\bf 0, \bf I)$.

The reverse process is trained to match the actual time reversal of the forward process. 
The loss takes the form of 
\begin{equation*}
    \mathbb E_{k \sim \mathcal U [\![1, K]\!], \vs^0 \sim p_{\text{data}}(\vs^0), \vepsilon \sim  \mathcal N(\bf 0, \bf I)}\left[\lVert \vepsilon - \vepsilon_\theta(\vs^k, k)\rVert^2 \right]
\end{equation*}
where $\vs^k$ is computed from $\vs^0$ and $\vepsilon$, $\mathcal U [\![1, K]\!]$ is the discrete uniform distribution over $\{1, 2, \ldots, K\}$, and $\vepsilon_\theta$ is a deep neural network with parameters $\theta$ that predicts the noise $\vepsilon$ given $\vs^k$ and $k$.



\subsection{Action Prediction}
Given the current state $s_t$, we first predict a horizon of future states $\hat\vs \sim  p_{\theta}(\vs|s_t)$.
In the second phase, the objective is to determine the action $a_{t}$ that transitions $s_t$ to the next predicted state $\hat s_{t+1}$ in the sequence~$\hat\vs$.
To achieve this, we introduce an action model $f_\phi(s_t, s_{t+1}) = a_t$. 
The learning objective of $f_\phi$, detailed in section \ref{sec:deepensembles}, is to accurately predict $a_t$.

The overall generative modeling policy is the composition of the generative and action models.

\begin{algorithm}[tbh!]
    \caption{Uncertainty-Based Adaptive Planning} \label{alg:adapolicy}
    \begin{algorithmic}[1]
        \Require Generative distribution of states $p_\theta$, ensemble of action models $E = \{f_{\phi_1}, f_{\phi_2}, \ldots, f_{\phi_M}\}$, uncertainty threshold $\delta$
        \State $t \leftarrow 0$
        \State Observe initial state state $s_0$
        \While{episode is not done}
            \State Predict futrue states $\hat\vs \sim  p_{\theta}(\vs|s_t)$
            \State Predict action $a_t$ and uncertainty $u_t$  with $E$ (Eqs~\ref{eq:ens_at},~\ref{eq:ens_ut})
            \State Execute $a_t$, increment $t$, and observe new state $s_t$
            \State Update the state $\hat s_1 \leftarrow s_t$
            \State Predict $a_t$ and $u_t$ with $E$ 
            \State $i \leftarrow 1$
            \While{$i < H -1$ and $u_t < \delta$}
                \State Execute $a_t$, increment $t$, and observe new state~$s_t$
                \State Update the state $\hat s_i \leftarrow s_t$
                \State Predict $a_t$ and $u_t$ with $E$
                \State $i \leftarrow i + 1$
            \EndWhile
        \EndWhile
    \end{algorithmic}
\end{algorithm}
\section{Adaptive Decision-Making under Uncertainty}\label{sec:adaptive}
\subsection{Adaptive Policy}
Planning with generative models, though effective, incurs significant computational costs when performed at every iteration of the control loop. 
This is especially true for diffusion models because they require numerous passes through the underlying deep neural network to gradually produce data from random noise.
In contrast, action models, designed with simpler architectures, require considerably less computational effort. 
This insight leads to a strategy whereby the agent leverages the generative model's capacity to predict a long horizon of future states to execute multiple actions consecutively. 

This strategy naturally leads to a critical question: When should the agent update its planned trajectory of states or choose to execute an action?
We propose to use the uncertainty of the action model's predictions as a criterion for this process. 
To this end, the action model $f_\phi$ is modified to not only predict the next action, but also to estimate the uncertainty of its prediction.
This approach hinges on the premise that higher uncertainty signals a greater necessity for re-evaluation of the plan through the generative model, ensuring that subsequent decisions are made with the most current observations of the environment.

Specifically, we introduce the following adaptive policy. 
Starting from the current state, we generate a trajectory of states using the generative model and compute the initial action using the action model, which is then executed.
We then continue to predict and execute actions, adjusting the plan with each new observation, as long as the uncertainty remains below a predefined threshold.
The policy is detailed in Algorithm \ref{alg:adapolicy}.

The threshold is a tunable test-time hyperparameter that balances the trade-off between computational efficiency and safety. 
This flexibility allows users to adjust the threshold according to their specific needs without the necessity for retraining, enabling a single model to adapt to varying demands on computational resources and accuracy levels.

\subsection{Deep Ensembles for Predictive Uncertainty Estimation}\label{sec:deepensembles}
We implement a Deep Ensemble \cite{lakshminarayanan2017deepensembles} of action models to estimate the predictive uncertainty, which combines both aleatoric and epistemic uncertainties. 
Aleatoric uncertainty, which arises from inherent noise in the data, is quantified using the model’s output variance, while epistemic uncertainty, stemming from the model's lack of knowledge, is captured by the variability among the different models in the ensemble.

Deep ensembles are straightforward to implement and require minimal or no modifications to a standard action model’s architecture, which in our case is a neural network with parameters $\phi$. 

Let $x$ denote the input features to the model, and $y$ the actual action observed in the data, against which the model’s predictions are compared.
The network’s final layer outputs two values:
$$f_\phi(x) = \left(\mu_\phi(x), \sigma_\phi^2(x)\right),$$
where the predicted mean $\mu_\phi(x)$ represents the model’s expectation of the output and the variance $\sigma_\phi^2(x)$ quantifies the model’s aleatoric uncertainty in its predictions. 
The model is trained by minimizing the Negative Log-Likelihood~(NLL):
\begin{equation}\label{eq:NLL}
-\log p_\phi(y|x) = \frac{1}{2} \log(\sigma_\phi^2(x)) + \frac{(y - \mu_\phi(x))^2}{2\sigma_\phi^2(x)}.
\end{equation}
The first term, penalizes large variances.
The second term, is essentially a scaled mean squared error that becomes more penalizing when the variance is small but the prediction error is large.
This modeling approach assigns higher variance for inputs where the model predicts outcomes with less certainty.

To ensure the variance $\sigma_\phi^2(x)$ remains positive, we apply the softplus function to the network's variance output: $\log(1 + \exp(\cdot))$, and introduce a minimum variance of $10^{-6}$ for numerical stability.

Each model in the ensemble is trained on the entire dataset but initialized with random parameters to introduce diversity in the predictions. 
The action at time $t$ is determined by averaging the mean predictions across all $M$ ensemble members:
\begin{equation}\label{eq:ens_at}
a_t = \frac{1}{M} \sum_{m=1}^{M} \mu_{\phi_m}(x).
\end{equation}

The predictive uncertainty, capturing both aleatoric and epistemic uncertainties, is computed as the sum of the models' average variance and the variance of the ensemble's mean predictions:
\begin{equation}\label{eq:ens_ut}
u_t = \frac{1}{M} \sum_{m=1}^{M} \sigma^2_{\phi_m}(x) + \text{Var}\left(\{\mu_{\phi_m}(x)\}_{m=1}^{M}\right).
\end{equation}
This approach gives us a principled way to quantify the model's uncertainty in its action predictions.

\section{Experiments}\label{sec:experiments}
\begin{table*}[tbh!]
\centering
\label{my-label}
\begin{tabular}{@{}c c c c c c c c c c@{}}
\toprule
\textbf{Dataset} & \textbf{Env.} & \multicolumn{2}{c}{\textbf{DD}} & \multicolumn{2}{c}{\textbf{Static Plan Execution}} & \multicolumn{2}{c}{\textbf{EA MSE (Ours)}} & \multicolumn{2}{c}{\textbf{EA NLL (Ours)}} \\
\cmidrule(lr){3-4} \cmidrule(lr){5-6} \cmidrule(lr){7-8} \cmidrule(lr){9-10}
 &  & \makecell{Return} & \makecell{Saved NFE} & \makecell{Return} & \makecell{Saved NFE} & \makecell{Return} & \makecell{Saved NFE} & \makecell{Return} & \makecell{Saved NFE} \\
\midrule
Medium & Hopper & 49.9 & 0\% & 5.3 & 98.7\% & 54.1 & 85.3\% & 62.1 & 91.1\% \\
Medium & Walker & 74.5 & 0\% & 3.9 & 98.7\% & 74.8 & 92.1\% & 52.5 & 76.8\% \\
\midrule
Med-Rep & Hopper & 59.8 & 0\% & 7.1 & 98.7\% & 72.0 & 69.0\% & 69.7 & 17.0\% \\
Med-Rep & Walker & 62.7 & 0\% & 13.3 & 98.7\% & 66.4 & 91.3\% & 62.8 & 90.6\% \\
\midrule
Med-Exp & Hopper & 110.0 & 0\% & 57.3 & 98.9\% & 109.0 & 84.6\% & 109.1 & 93.0\% \\
Med-Exp & Walker & 78.8 & 0\% & 15.6 & 98.8\% & 83.1 & 96.9\% & 80.4 & 89.7\% \\
\bottomrule
\end{tabular}
\caption{\textit{This table presents the average normalized rewards achieved by Decision Diffuser (DD), Static Plan Execution, and Ensemble Action (EA) for both MSE and NLL training criteria. 
The table also reports the percentage of actions executed without sampling from the generative model.}}
\label{tab:results_summary}
\end{table*}

\subsection{Experimental Setup}
This section evaluates the efficacy of our proposed adaptive policy, which we name Ensemble Action, in offline reinforcement learning (RL) control tasks. We utilize the D4RL Hopper and Walker locomotion environments with different dataset settings:
\begin{itemize}
	\item \textbf{Medium}: Generated from 1 million timesteps by a medium policy, achieving approximately one-third of an expert policy's score.
	\item \textbf{Medium-Replay}: Includes the replay buffer from an agent trained to a medium policy's performance level.
	\item \textbf{Medium-Expert}: Combines 1 million timesteps from the medium policy with an additional 1 million timesteps from an expert policy.
\end{itemize}
We assess Ensemble Action's performance in terms of time and accuracy. 
Our analysis focuses on potential reductions in network function evaluations (NFEs) \cite{karras2022edm} required by the diffusion model. 
Table \ref{tab:results_summary} demonstrates the trade-off between the average normalized reward \cite{fu2020d4rl} and the percentage of saved NFEs, compared to a policy that samples from the diffusion model at every timestep.

Given its foundational role in decision-making with generative models, Decision Diffuser \cite{ajay2022decisiondiffuser} serves as our primary benchmark. 
Sampling trajectories with Decision Diffuser involves 100 denoising iterations using DDPM whenever the agent take an action, significantly increasing the planning time.
Additionally, we compare our approach to Static Plan Execution, which commits to a predetermined sequence of actions for the entire horizon without replanning. 
This method, while computationally efficient due to the absence of ongoing planning, assumes the initial plan remains optimal throughout its execution, potentially limiting effectiveness in dynamic or unpredictable conditions.

In our experiments, we set the planning horizon ($H$) to 100. 
We use identical diffusion models for state prediction in both Ensemble Action and Decision Diffuser. 
The action model is a simple 2-layer perceptron, with each layer consisting of 512 units.
We create ensembles of 5 action models and conduct 50 random simulations for each task. 
We selected different uncertainty thresholds ($\delta$) for each dataset, choosing values that achieved a favorable balance between prediction accuracy and time.

As anticipated, Static Plan Execution shows lower return, likely due to compounded errors from individual actions, underscoring the importance of adaptability in dynamic settings. 
In contrast, Ensemble Action maintains comparable rewards to the Decision Diffuser baseline while significantly reducing the need for network function evaluations by up to 93\%.
However, an outlier in this trend is observed in the Medium-Replay Hopper scenario for the NLL-trained Ensemble Action, where only 17\% of NFEs were saved. 
This variability emphasizes the need for adaptive thresholds and possibly fine-tuning the decision criteria based on the characteristics of each dataset to optimize both performance and efficiency.

We also evaluate the computational efficiency of generating 100 steps on a Tesla V100-PCIe-32GB GPU by comparing the generation times of an ensemble of action models against those of Decision Diffuser and Decision Transformer \cite{chen2021decisiontransformer}. 
The results, presented in Table \ref{tab:time}, indicate that the ensemble of action models achieves the fastest step generation times, recording 0.13 seconds for the Hopper and 0.16 seconds for the Walker environments.
Decision Diffuser required significantly more time, approximately 20.19 seconds for Hopper and 20.81 seconds for Walker, demonstrating the superior efficiency of the ensemble approach in rapid step generation.

\begin{figure}[b!]
\centering
\begin{tikzpicture}
\begin{axis}[
    xlabel={Steps},
    ylabel={Time (seconds, log scale)},
    xmin=0, xmax=1000,
    ymin=1, ymax=1500,
    xtick={0,200,400,600,800,1000},
    ytick={1, 10, 100, 1000},
    ymode=log,
    log basis y={10},
    legend pos=south east,
    ymajorgrids=true,
    grid style=dashed,
]

\addplot[
    color=blue,
    no marks,
    line width=1pt,
    ]
    coordinates {
    (0,1.5469977855682373)(10,1.6110901832580566)(20,1.6728665828704834)(30,1.7333064079284668)(40,1.791874885559082)(50,1.8529102802276611)(60,1.9146080017089844)(70,1.973449468612671)(80,2.032578706741333)(90,2.0915679931640625)(100,3.5246551036834717)(110,3.585972785949707)(120,3.645848274230957)(130,3.7062060832977295)(140,5.136369943618774)(150,5.196467876434326)(160,5.256012678146362)(170,5.317269802093506)(180,5.38025426864624)(190,5.442133188247681)(200,6.878528833389282)(210,6.937645673751831)(220,6.998430252075195)(230,7.061397552490234)(240,7.121417999267578)(250,7.18050742149353)(260,7.238763093948364)(270,7.298725843429565)(280,7.36003303527832)(290,8.79665493965149)(300,8.855530500411987)(310,8.916494131088257)(320,8.97559404373169)(330,9.038690328598022)(340,9.101417303085327)(350,9.160306692123413)(360,9.21924877166748)(370,9.277436971664429)(380,9.337220907211304)(390,10.760625123977661)(400,10.823070764541626)(410,10.883571863174438)(420,10.943957328796387)(430,11.006796836853027)(440,11.070441246032715)(450,12.501249074935913)(460,12.562077283859253)(470,12.625581979751587)(480,12.686493396759033)(490,12.74606466293335)(500,12.806505680084229)(510,12.867632150650024)(520,12.927109479904175)(530,12.985427856445312)(540,13.04460072517395)(550,14.462339639663696)(560,14.521734952926636)(570,14.58173394203186)(580,14.641868352890015)(590,14.701544284820557)(600,14.759921073913574)(610,14.819701671600342)(620,14.87862229347229)(630,14.937837839126587)(640,14.99856972694397)(650,16.422388315200806)(660,16.48196816444397)(670,16.540457010269165)(680,16.600705862045288)(690,16.65984344482422)(700,16.719855785369873)(710,16.783989667892456)(720,16.84488081932068)(730,16.905431747436523)(740,16.964150428771973)(750,18.387532711029053)(760,18.446609497070312)(770,18.505927562713623)(780,18.567891597747803)(790,18.62886071205139)(800,18.689674139022827)(810,18.750879049301147)(820,18.811400651931763)(830,18.872573375701904)(840,18.933314323425293)(850,20.354904651641846)(860,20.414449453353882)(870,20.475852966308594)(880,20.53616499900818)(890,20.595974922180176)(900,20.655983448028564)(910,22.083903074264526)(920,22.142943382263184)(930,22.201972007751465)(940,23.638062953948975)(950,23.696776628494263)(960,23.75698447227478)(970,23.816001176834106)(980,23.876216411590576)(990,23.936732292175293)
    };
    \legend{Ensemble Action}

\addplot[
    color=red,
    no marks,
    line width=1pt,
    ]
    coordinates {
    (0,1.573960304260254)(10,15.653045177459717)(20,29.70068049430847)(30,43.75822114944458)(40,57.836050510406494)(50,71.9432737827301)(60,86.0384292602539)(70,100.12590003013611)(80,114.19235730171204)(90,128.254380941391)(100,142.36342358589172)(110,156.45982670783997)(120,170.55369639396667)(130,184.6305115222931)(140,198.70955324172974)(150,212.77156519889832)(160,226.82326650619507)(170,240.87623023986816)(180,254.9258005619049)(190,268.99321150779724)(200,283.06316781044006)(210,297.15159606933594)(220,311.22128653526306)(230,325.30102586746216)(240,339.3675968647003)(250,353.4235715866089)(260,367.5026054382324)(270,381.5399191379547)(280,395.6095449924469)(290,409.67500495910645)(300,423.7223744392395)(310,437.7630846500397)(320,451.8151230812073)(330,465.8724732398987)(340,479.927205324173)(350,493.9875741004944)(360,508.00497102737427)(370,522.0416197776794)(380,536.0840535163879)(390,550.0981888771057)(400,564.1358714103699)(410,578.2036304473877)(420,592.2481534481049)(430,606.2942914962769)(440,620.3226313591003)(450,634.3591732978821)(460,648.4294290542603)(470,662.4786348342896)(480,676.5204396247864)(490,690.5599608421326)(500,704.5964307785034)(510,718.6744890213013)(520,732.6981835365295)(530,746.7231142520905)(540,760.7317776679993)(550,774.7505254745483)(560,788.7626519203186)(570,802.7877020835876)(580,816.8299655914307)(590,830.8652830123901)(600,844.9001007080078)(610,858.9167470932007)(620,872.9823024272919)(630,887.0024828910828)(640,901.020293712616)(650,915.099778175354)(660,929.1612012386322)(670,943.1899721622467)(680,957.209742307663)(690,971.2579035758972)(700,985.2790911197662)(710,999.3315417766571)(720,1013.3554990291595)(730,1027.3951165676117)(740,1041.4559967517853)(750,1055.4865288734436)(760,1069.5262546539307)(770,1083.5487632751465)(780,1097.5825581550598)(790,1111.642471075058)(800,1125.6942989826202)(810,1139.7597572803497)(820,1153.802615404129)(830,1167.8564686775208)(840,1182.3791363239288)(850,1196.4199657440186)(860,1210.4577083587646)(870,1224.508395910263)(880,1238.5588405132294)(890,1252.6031589508057)(900,1266.6530871391296)(910,1280.7187163829803)(920,1294.757571220398)(930,1308.8057193756104)(940,1322.8768360614777)(950,1336.9694623947144)(960,1351.0436050891876)(970,1365.1099050045013)(980,1379.159496307373)(990,1393.2082636356354)
    };
    \addlegendentry{Decision Diffuser}
\end{axis}
\end{tikzpicture}
\caption{\textit{Ensemble Action completes 1000 steps in under 25 seconds, while the Decision Diffuser takes over 23 minutes, resulting in a 55x speedup.}}
\label{fig:steps_vs_time}
\end{figure}
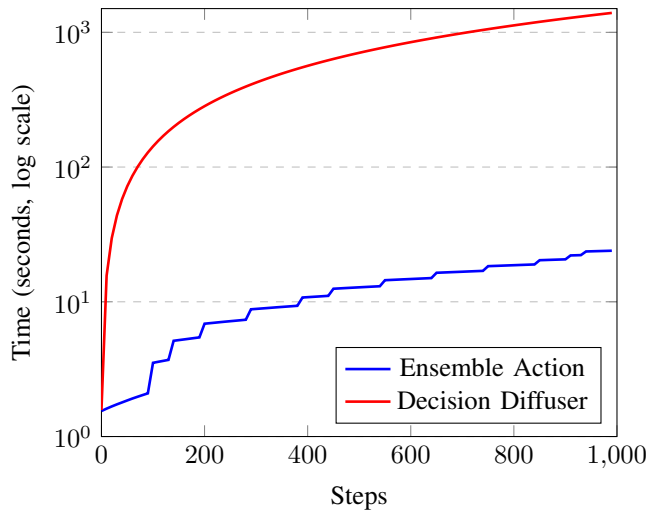

The plot in Figure \ref{fig:steps_vs_time} illustrates the time required to complete a 1000-step episode using Hopper environment for both Ensemble Action and Decision Diffuser. 
It is evident that our adaptive policy significantly outperforms Decision Diffuser in terms of computational efficiency. 
Specifically, our adaptive policy completes the 1000 steps in less than 25 seconds, while Decision Diffuser takes over 23 minutes. 
This difference highlights the efficiency of our approach in rapidly processing steps, thereby enabling quicker decision-making.
\begin{table}[bth]
\centering
\begin{tabular}{lccc}
\hline
\textbf{Env}     & \textbf{DT} & \textbf{DD} & \textbf{EA (Ours)} \\ \hline
Hopper  & 0.33                  & 20.19             & 0.13                      \\
Walker  & 0.34                  & 20.81             & 0.16                      \\ \hline
\end{tabular}
\caption{\textit{Average time to generate 100 steps in seconds for Decision Transformer (DT), Decision Diffuser (DD), and Ensemble Action (EA)}}
\label{tab:time}
\end{table}

\subsection{Ablation Study}
\begin{table*}[tbh!]
\centering
\begin{tabular}{@{}c c c c c c c@{}}
\toprule
\textbf{Dataset} & \textbf{Environment} & \multicolumn{5}{c}{\textbf{Number of Ensemble Members} \(M\)} \\
\cmidrule{3-7} 
 & & \(M=1\) & \(M=2\) & \(M=3\) & \(M=4\) & \(M=5\) \\
\midrule
Medium-Expert & Hopper & 110.0 (0\%) & 106.6 (78.2\%) & 102.1 (97.7\%) & 107.7 (98.1\%) & 109.0 (84.6\%) \\
Medium-Expert & Walker & 78.8 (0\%) & 79.5 (89.1\%) & 78.3 (97.2\%) & 75.2 (96.9\%) & 83.1 (96.9\%) \\
\bottomrule
\end{tabular}
\caption{\textit{Impact of the Number of Ensemble Members \(M\) on the Ensemble Action Performance. This table displays the average normalized reward and the percentage of saved NFEs.}}
\label{tab:ensembleMembersImpact}
\end{table*}

Although training with NLL directly estimates aleatoric uncertainty by learning an additional output for variance, this introduces complexity and may not be necessary in environments with deterministic state transition functions. 
In contrast, while MSE loss does not inherently quantify aleatoric uncertainty, training action models with MSE is simpler and does not require any alterations to the original architecture or training procedures of the action model used in Decision Diffuser. 
In that case, the action model directly estimates the next action
$$f_\phi(x) = a_t$$
and the training objective is
\begin{equation*}\label{eq:MSE}
    \mathbb E_{a_t \in \mathcal D}\left[ \lVert a_t - f_\phi(x)\rVert^2\right].
\end{equation*}
The total uncertainty in Equation \ref{eq:ens_ut} is replaced by the epistemic uncertainty alone:
\begin{equation*}\label{eq:ens_ut_mse}
u_t = \text{Var}\left(\{f_{\phi_m}(x)\}_{m=1}^{M}\right).
\end{equation*}
We present the results of the Ensemble Action trained with MSE in Table \ref{tab:results_summary}.
In our experiments, we find that training with MSE yields results comparable to those obtained with NLL, while being simpler.

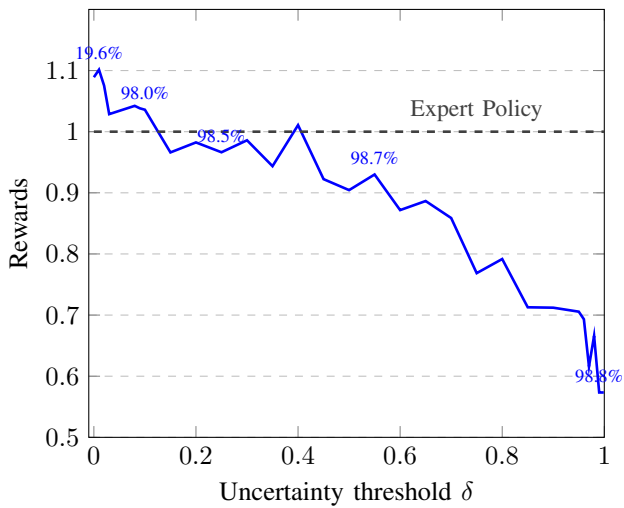
\begin{figure}[tbh!]
\centering
\begin{tikzpicture}
\begin{axis}[
    xlabel={Uncertainty threshold $\delta$},
    ylabel={Rewards},
    xmin=-0.01, xmax=1,
    ymin=0.5, ymax=1.2,
    xtick={0, 0.2, 0.4, 0.6, 0.8, 1.0},
    ytick={0.5, 0.6, 0.7, 0.8, 0.9, 1.0, 1.1},
    legend pos=north east,
    ymajorgrids=true,
    grid style=dashed,
    xticklabel style={rotate=0, /pgf/number format/fixed},
    point meta=explicit symbolic, 
    nodes near coords, 
    nodes near coords align={vertical}, 
    every node near coord/.append style={font=\fontsize{7}{9.6}\selectfont, color=blue} 
]

\addplot+[color=blue, no marks, nodes near coords, line width=1pt,] coordinates {
    (0., 1.08908591)
    (0.01, 1.10139728) [19.6\%]
    (0.02, 1.07540401)
    (0.03, 1.02870623) 
    (0.08, 1.04225661) 
    (0.09, 1.03826469)
    (0.1, 1.03589035) [98.0\%]
    (0.15, 0.96608851)
    (0.2, 0.98243114)
    (0.25, 0.96631816) [98.5\%]
    (0.3, 0.98593197)
    (0.35, 0.94358074)
    (0.4, 1.0107319)
    (0.45, 0.92236766)
    (0.5, 0.90440156)
    (0.55, 0.92998245) [98.7\%]
    (0.6, 0.87173005)
    (0.65, 0.88655655)
    (0.7, 0.85877148)
    (0.75, 0.7685842)
    (0.8, 0.79172088)
    (0.85, 0.71274144)
    (0.9, 0.71208978)
    (0.95, 0.7055813)
    (0.96, 0.69320287) 
    (0.97, 0.61583186) 
    (0.98, 0.66597977) 
    (0.99, 0.57338374) [98.8\%]
    (0.999, 0.57338374)
};

\draw[dashed, darkgray, line width=1pt] (axis cs:0,1) -- (axis cs:1,1) node[pos=0.75, above, font=\small] {Expert Policy};
\end{axis}
\end{tikzpicture}
\caption{\textit{Impact of varying $\delta$ on rewards for the Ensemble Action model in the Hopper Medium-Expert dataset, with specific $\delta$ values showing saved NFEs.}}
\label{fig:reward_vs_delta}
\end{figure}

We also explored the impact of varying the size of the ensemble within the Hopper and Walker environments, using the Medium-Expert dataset. 
In these experiments, we train the ensemble members with MSE loss. 
Our findings indicate that reducing the ensemble size to four or three action models still yields robust performance. 
This reduction decreases computational overhead at test time and reduces training time, as fewer models require less time to train. 
Notably, when the ensemble size is reduced to one ($M=1$), the Ensemble Action method essentially replicates the Decision Diffuser algorithm, which explains why the saved NFEs for $M=1$ are 0\%. 
The detailed outcomes of these experiments are presented in Table \ref{tab:ensembleMembersImpact}.

Furthermore, we analyzed how the uncertainty threshold influences replanning decisions and overall reward outcomes, focusing on the Hopper Medium-Expert environment.
Our findings, illustrated in Figure \ref{fig:reward_vs_delta}, demonstrate that fine-tuning the uncertainty threshold allows us to significantly reduce the need for NFEs by up to 98\%, while maintaining expert-level rewards. 
Additionally, we observed, as expected, that the rewards increase as the uncertainty threshold decreases.
These results underscore the potential of our approach to dramatically enhance computational efficiency without compromising the quality of decision-making.

\section{Discussion and Conclusion}\label{sec:discusionconclusion}
In this study, we introduced an adaptive policy aimed at reducing planning time when using generative models. Our policy first produces a trajectory of future states with the generative model. Then, a Deep Ensemble of action models interprets this trajectory to predict the next actions, continuing to do so as long as its uncertainty stays below a threshold. This mechanism enables the policy to adaptively determine when to replan, optimizing computational efficiency while ensuring decisions are based on reliable predictions.

This adaptive strategy significantly reduces the need for frequent calls to the generative model--approximately 90\% fewer calls in most of our experiments--without sacrificing decision quality.
However, there are limitations and areas for future research that should be addressed. 
Exploring the applicability and scalability of our approach to more complex scenarios, such as real-world robotics and autonomous driving, presents exciting challenges. 
Additionally, it would be interesting to compare the computational effort of a non-generative state-of-the-art offline RL method, such as an actor-critic method, with our proposed method. 
Further research could also explore this strategy with other types of policies or generative models that decouple the state and action prediction phases.

In conclusion, our research lays the groundwork for more efficient use of generative models in decision-making, suggesting a path toward real-time decision-making systems.




\bibliographystyle{IEEEtran}
\bibliography{IEEEabrv, references}

\end{document}